\documentclass[a4paper,12pt]{article}
\usepackage{amsthm}
\usepackage{amssymb}
\usepackage{amsmath}
\usepackage{amsfonts}
\usepackage{color}
\usepackage{graphicx}
\usepackage[square, comma, sort&compress, numbers]{natbib}
\date{}

\newtheorem{remark}{Remark}[section]

\newtheorem{theorem}{Theorem}[section]
\newtheorem{proposition}{Proposition}[section]
\newtheorem{corollary}{Corollary}[section]

\def\b1{\mbox{\boldmath $1$}}

\newenvironment{demo*}{\vspace{3mm}\noindent{\bf Proof.}}{\hfill $\Box$ \vspace{3mm}}

\begin{document}
\title{\bf \Large {Elliptical Distributions-Based Weights-Determining Method for  OWA Operators}}
{\color{red}{\author{\normalsize{Xiuyan Sha}\\{\normalsize\it  School of Statistics, Qufu Normal University}\\\noindent{\normalsize\it Shandong 273165, China}\\e-mail:   shaxiuyan@sina.com\\
\normalsize{Zeshui Xu}\\{\normalsize\it Business School,  Sichuan University}\\
\noindent{\normalsize\it Chengdu 610064, China}\\
e-mail: xuzeshui@263.net\\
\normalsize{Chuancun Yin}\thanks{Corresponding author.}\\{\normalsize\it   School of Statistics, Qufu Normal University}\\
\noindent{\normalsize\it Qufu 273165, China}\\
e-mail: ccyin@mail.qfnu.edu.cn\\}}}
\maketitle
\vskip0.01cm
\noindent{\large {\bf Abstract:}}  {The ordered weighted averaging
(OWA) operators  play a crucial role in aggregating multiple criteria evaluations
into an overall assessment supporting the decision makers' choice.   One key point steps is to determine the
associated weights.   In this paper, we first briefly review some main methods for determining
the weights by using distribution functions. Then we propose a new approach
 for determining OWA weights by using the RIM quantifier. Motivated by the idea of  normal distribution-based method
to determine the OWA weights,  we  develop a  method based on elliptical distributions for determining
the OWA weights, and some of its desirable properties have been
investigated. }




\numberwithin{equation}{section}
\section{Introduction}\label{intro}
 Aggregation operators play important roles in the theory of fuzzy sets. The ordered weighted averaging operator (OWA) introduced by
Yager (1988) is a fundamental aggregation operator which generalizes {\sl or} and {\sl and} aggregation operators. The OWA operators have been applied in diverse fields such as  multicriteria and group decision making (Herrera (1995,1996), Liu et al. (2018), Yager and Alajlan (2018)),
data mining (Torra, 2004),  asset management and actuarial science (Casanovas et al. 2016, Merig\'o, 2018) and approximate reasoning (Dujmovic, 2006).
The {\sl orness/andness} measure of an OWA operator plays an important role in  the studies of the OWA operators in both theoretical and applied areas; see  Ahn (2006), Filev and  Yager (1998),  Jin (2015),
Liu and  Chen (2004), Liu and Han (2008),
 Jin and Qian (2015), Xu (2005),  Yager et al. (2011), Reimann et al. (2017).  Orness/andness measure reflects the extent of orlike/andlike of the aggregation result under an OWA operator.

The OWA operator provides a parameterized class of mean type operators which can be used to aggregate a collection of arguments. The parameterization is accomplished by the choice of the characterizing OWA weights that are multiplied by the argument values in a linear type aggregation.   One important issue is the determination of the associated weights of the  operator. During the last two decades, scholars proposed a large variety of weights determination methods for the OWA operators with numerous different constraints, such as  maximizing deviation method (Wei and Feng (1998))and  Gaussian distribution-based
method (Xu (2005)), Sadiq and Tesfamariam (2007) extended this method by using
the probability density functions, see also  Lenormand (2018) for truncated
distributions method. 
Liu (2007) proved the solution equivalence of the minimum
variance problem and the minimax disparity problem.
Wang and Xu (2008) introduced a method utilizing the binomial distribution for obtaining the OWA operator weighting vector.
Wang et al. (2007) proposed
least squares deviation and Chi-square models to produce the OWA weights
with a given orness degree.  Liu (2008) gave a more general form of the OWA
operator determination methods with a convex objective function, which can include
the maximum entropy and minimum variance problems as special cases.
Merig\'o (2012a, 2012b) presented the probabilistic weighted average  operator and the
probabilistic OWA operator, respectively. The method of Lagrange multipliers to determine the optimal weighting vector was proposed by Sang and Liu (2014). Note also that there are several generalizations of the OWA operators, such as
weighted OWA  operator (Yager, 1993), induced OWA  operator (Yager, 1996), generalized OWA operator
(Beliakov, 2005), probabilistic
OWA (POWA) operator (Merig\'o, 2012), and so on. Jin and Qian (2016) proposed a new tool called OWA generation function which is inspired from the generating function in probability theory.   Recent extensions on OWA can be found in Yager (2017),  Mesiar et al. (2018), Wang et al. (2018) and Yager and Alajlan (2018), among others.

In this paper, we first give a survey of the existing main methods and
then develop two novel practical methods based on the quantifier functions and  elliptical distributions for determining
the OWA weights.

The rest of the paper is structured as follows:  In Section 2 , we give a brief overview of the existing OWA literature. In Section 3, we propose  a new method to determine  weights by the quantifier function and compare the properties with usual weights determined by the same quantifier function. We describe   an elliptical distribution-based method to determine the OWA weights in Section 4,  which can be seen as the extension of Gaussian-based method in  Xu (2005). The final section provides a summary and concludes the paper.

\section{ Review of the OWA operators}

In this section we  review the  notion  and some facts about the  OWA operator and its generalization.
The ordered weighted averaging (OWA)  operator was introduced by Yager (1988)
 which is a generation of arithmetic mean, maximum and minimum
operators. It provides a parameterized family of
aggregation operators which can be defined as follows:

\noindent {\bf Definition 2.1.}  (Yager (1988))\; An OWA operator of dimensions $n$ is a mapping
OWA: $\Bbb{R}^n\rightarrow\Bbb{R}$
such that
$$OWA(a_1,\cdots,a_n)=\sum_{i=1}^n w_i a_{(i)},$$
where  $(\cdot)$ is a permutation of $\{1, \cdots, n\}$ such that $a_{(1)}\ge a_{(2)}\ge \cdots\ge a_{(n)}$, i.e.,  $a_{(j)}$ is the $j$th largest element of the collection of the aggregated objects
$a_1, a_2,\cdots, a_n$ and  the $w_i$'s are weights satisfying $w_i\in [0,1]$ and $\sum_{i=1}^n w_i=1$.

Each weight $w_i$ is associated with the ordered
position $i$ rather than the argument $a_i$.  Yager (1988)  defined two important measures
associated with an OWA operator. The first
measure, called the dispersion (or entropy) of an
OWA vector $W$ is defined as:
$$Disp(W)=-\sum_{i=1}^n w_i\ln w_i.$$
 The second measure called ``orness"  is defined as:
\begin{equation}
orness(W)=\sum_{j=1}^n\frac{n-j}{n-1}w_j.
\end{equation}

The measure of ``andness" associated with an OWA operator is the complement of its ``orness", which is defined as
$andness(W)=1-orness(W).$
It is noted that different OWA operators are
distinguished by their weighting function.  Yager (1993) discussed various different examples of weighting vectors. For example,
 $W^*=(1,0,\cdots,0), W_*=(0,\cdots,0,1)$ and $W_A=(\frac{1}{n},\cdots,\frac{1}{n})$, which
correspond to the $max$, $min$,  and $ mean$. Obviously,  $orness(W^*)=1$,  $orness(W_*)=0$ and   $orness(W_A)=\frac12$.
 It has been established in Yager (1988) that the OWA operator has the
following properties: \\
Commutativity:  $OWA(a_1, \cdots, a_n)=OWA(\pi(a_1), \cdots, \pi(a_n))$ for any permutation $\pi$; \\
 Monotonicity:  $OWA(a_1, \cdots, a_n)\ge  OWA(b_1, \cdots, b_n)$ if $a_i\ge b_i$  for each $i$;\\
 Boundedness: $\min_i\{a_i\}\le OWA(a_1, \cdots, a_n)\le \max_i\{a_i\}$;\\
Idempotency:  OWA$(a,\cdots,a)=a$.

A very useful approach for obtaining the OWA weights is the functional method introduced by Yager (1996).
A fuzzy subset $Q$ of the real line is called a Regular Increasing
Monotone (RIM) quantifier if $Q(0)=0, Q(1)=1$ and $Q(r_1)\ge Q(r_2)$ if $r_1>r_2$.
We shall call a quantifier $Q$ regular unimodal (RUM)
if $Q(0)=Q(1)=0$; $Q(r)=$l  for  some $a\le r\le b$; there exist two values $r_1$  and $r_2$ such that
if $a\le r_2\le r_1$ then $Q(r_2)\le Q(r_1)$; if $r_2\le r_1\le b$ then $Q(r_2)\ge Q(r_1)$.   These
functions were also denoted as basic unit-interval monotonic (BUM)
functions in Yager (2004). Examples of this kind of quantifier are $all$, $most$, $many$, $there$ $exists$ (Yager, 1996).
The quantifier $all$ is represented by the fuzzy subset $Q_*(x)={\bf 1}_{\{x=1\}}$, the quantifier $there\; exists$, $not\; none$, is defined as
$Q^*(x)={\bf 1}_{\{x\neq 0\}}$, where ${\bf 1}_{A} = 1$ if the event $A$ is true, or $0$ otherwise.

Given a RIM quantifier,  Yager (1988) generated  the OWA weights by
$${{w}}_i=Q\left(\frac{i}{n}\right)-Q\left(\frac{i-1}{n}\right),\; i=1, 2,\cdots, n.$$
The quantifier guided aggregation with the OWA operator is
$$OWA_{{Q}}(a_1,\cdots,a_n)=\sum_{i=1}^n { w}_i a_{(i)},$$
where  $a_{(j)}$ is the $j$th largest element of the collection of the aggregated objects
$a_1, a_2,\cdots, a_n$. We refer the reader to Yager (1988) for the definition and more examples.

Yager (1996)  extended the orness measure of the OWA operator (2.1),
and defined the orness of a RIM quantifier:
$$orness(Q)=\lim_{n\to\infty}\frac{1}{n-1}\sum_{j=1}^{n-1}Q\left(\frac{j}{n}\right)=\int_0^1 Q(r)dr.$$
 We have $orness(Q_*) = 0$, $orness(Q^*) = 1$, and $orness(Q_A)=\frac12$, where $Q_A(x)=x$.

Note that, sometimes, as in fuzzy set theory, it is
better to use in the definition a mapping OWA: ${[0,1]}^n\rightarrow {[0,1]}$. In this special case, the OWA also has the following properties
$OWA(0,\cdots,0)=0$ and $OWA(1,\cdots,1)=1$.

\noindent{\bf Definition 2.2.} (Yager, 1992) \ Assume that $W=(w_1,\cdots, w_n)$ is an OWA weighting vector of dimension
$n$. We shall say that $\widehat{W}=(\hat{w}_1,\cdots, \hat {w}_n)$ is its dual $\widehat{OWA}$ weighting vector if $\hat{w}_i=w_{n-i+1}$.
A weighting vector $W$ is said to be symmetric if  $w_i=w_{n-i+1}$.

The dual OWA operator    has the property $orness(\hat {W})+orness(W)=1$, and
in the special case when  $a_i\in [0, 1]$ (see also Yager and  Alajlan (2016)):
$$OWA_W(a_1,\cdots, a_n)=1-OWA_{\hat W}(1-a_1,\cdots, 1-a_n).$$

To establish the corresponding relationship between the OWA operator and the RIM quantifier, a generating function
representation of the RIM quantifier was proposed by Liu  (2005).
  A function $f$ on $[0,1]$ is called the generating function  of RIM quantifier $Q$, if it satisfies $f\ge 0$, $\int_0^1 f(x)dx=1$ and $Q(x)=\int_0^x f(z)dz.$ Obviously, $orness(Q)=1-\int_0^1 tf(t)dt$.
Here we will list some properties of the quantifier generating
function and the OWA operator:

\begin{proposition} (Liu and Han (2004))\; For the RIM quantifiers $Q$ and $G$ which
are determined by $f$ and $g$ respectively, if $g(x)=f(1-x)$,
then $orness(G)=1-orness(Q)$.
\end{proposition}

\begin{proposition} (Liu and Han (2008))\; For the  RIM quantifiers $Q$ and $G$,  $orness(Q)\ge orness(G)$ and $OWA_{{Q}}(a_1,\cdots,a_n)\ge OWA_{{G}}(a_1,\cdots,a_n)$ (for all $a_1,\cdots,a_n$) if and only if for
every rational point $x\in [0, 1], Q(x)\ge G(x)$.
\end{proposition}
A number of different methods have been suggested
for obtaining the weights associated with the OWA operator
(Xu, 2005). Moreover,
Xu (2005)  introduced a
procedure for generating the OWA weights based on
 the normal distribution (or Gaussian distribution).
Consider a normal distribution $N(\mu_n, \sigma_n^2)$, where
$$\mu_n=\frac{n+1}{2},\, \sigma_n^2=\frac{1}{n}\sum_{i=1}^n (i-\mu_n)^2=\frac{n^2-1}{12}.$$

The associated OWA weights is defined as:
$$w_i=\frac{e^{-(i-\mu_n)^2/2\sigma_n^2}}{\sum_{j=1}^n e^{-(j-\mu_n)^2/2\sigma_n^2}}, \; i=1,2,\cdots, n.$$

It is clear that $w_i\in [0,1]$ and $\sum_{i=1}^n w_i=1$ and $w_i$ is symmetric, that is, $w_i=w_{n+1-i}$.
 The prominent characteristic
of the developed method is that it can relieve the influence of unfair arguments
on the decision results by assigning low weights to those ``false" or ``biased"
ones.

\section{Determine  weights by the quantifier function}

A fuzzy subset $Q$ of the real line is called a Regular Increasing
Monotone (RIM) quantifier if $Q(0)=0, Q(1)=1$ and $Q(r_1)\ge Q(r_2)$ if $r_1>r_2$.
 RIM quantifiers   can be considered as a starting point for constructing families of the   OWA operators.
 Thus, constructions of  the RIM quantifiers  are very important  in producing various   OWA operators.  One way to generate  RIM quantifier   is  to use the method of mixing  along with finitely   RIM quantifiers  or  infinitely many   RIM quantifiers. Specifically,
if $Q_w$ ($w\in [a,b]$) is a one-parameter family of the   RIM quantifier, $\psi$ is an increasing function on $[a,b]$
  such that $\int_{[a,b]}d\psi(w)=1$, then the function
$Q(u)=\int_{[a,b]}Q_w(u) d\psi(w)$ is a   RIM quantifier.
In particular, if $\psi$ is discrete distribution, then   $Q(u)=\sum_i w_i Q_i(u)$ ($w_i\ge 0, \sum_i w_i=1)$.

The concept of dual OWA operators can be found in Yager (1992).
We now use  a RIM quantifier to give the definition of  dual OWA operators.

\noindent {\bf Definition 3.1.}  Let $Q$ be a RIM quantifier. An $\widetilde{OWA}_{Q}$ operator of dimension $n$ is a mapping
$\widetilde{OWA}_{Q}$: $\Bbb{R}^n\rightarrow\Bbb{R}$
such that
$$\widetilde{OWA}_{Q}(a_1,\cdots,a_n)=\sum_{i=1}^n {\tilde w}_i a_{(i)},$$
where  $a_{(j)}$ is the $j$th largest element of the collection of the aggregated objects
$a_1, a_2,\cdots, a_n$ and  the ${\tilde w}_i$'s are weights satisfying
$${\tilde w}_i=Q\left(1-\frac{i-1}{n}\right)-Q\left(1-\frac{i}{n}\right),\; i=1, 2,\cdots, n.$$

The orness of a RIM quantifier is defined as:
 $$\widetilde{orness}(Q)=\lim_{n\to\infty} \sum_{j=1}^n\frac{n-j}{n-1}{\tilde w}_j.$$
Note that ${\tilde w}_i\in [0,1]$ and $\sum_{i=1}^n {\tilde w}_i=1$. In addition,
\begin{equation}
\widetilde{orness}(Q)=\lim_{n\to\infty}\frac{1}{n-1}\sum_{i=1}^{n-1}\left(1-Q\left(1-\frac{i}{n}\right)\right)=1-\int_0^1 Q(r)dr.
\end{equation}
We have $\widetilde{orness}(Q_*) = 1$, $\widetilde{orness}(Q^*) = 0$, and $\widetilde{orness}(Q_A)=\frac12$, where $Q_A(x)=x$.
It is clear that ${\tilde w}_i=w_{n-i+1}$ and thus the $\widetilde{OWA}$ operator is just the dual OWA operator. Consequently,
  the $\widetilde{OWA}$ operator has the
following properties as the $OWA$ operator: Commutativity, Monotonicity, Boundedness and
Idempotency. In the special case, when the arguments $a_i\in[0, 1]$,  $\widetilde{OWA} (a_1,\cdots,a_n)\in [0, 1]$ and $\widetilde{OWA}$  also has the following properties $\widetilde{OWA} (0,\cdots,0)=0$
and $\widetilde{OWA} (1,\cdots,1)=1$. It is obvious that  if  the small aggregated objects have big weights  for the $OWA_Q$  operator,  then  the big aggregated objects have big weights  for the ${\widetilde{OWA}}_Q$  operator,  if  the big aggregated objects have big weights  for the $OWA_Q$  operator,  then  the small aggregated objects have big weights  for  the ${\widetilde{OWA}}_Q$  operator.

\noindent{\bf Example}\;  If $Q(u)=\sum_{i=1}^5\alpha_i Q_i(u)$, where $\alpha_i\ge 0$, $\sum_{i=1}^5\alpha_i=1$, $Q_1(u)=u$ and
\begin{eqnarray*}
 Q_2(u)=\left\{\begin{array}{ll} 0,  \ & {\rm if}\; u\in [0,1),\\
 1,\ &{\rm if}\;  u=1,
 \end{array}
  \right.
\end{eqnarray*}
\begin{eqnarray*}
 Q_3(u)=\left\{\begin{array}{ll} 0,  \ & {\rm if}\; u=0,\\
 1,\ &{\rm if}\;  u\in (0,1],
 \end{array}
  \right.
\end{eqnarray*}
\begin{eqnarray*}
 Q_4(u)=\left\{\begin{array}{ll} 0,  \ & {\rm if}\; u<\frac{n-k}{n},\\
 1,\ &{\rm if}\;  u\ge \frac{n-k}{n},
 \end{array}
  \right.
\end{eqnarray*}
and
\begin{eqnarray*}
 Q_5(u)=\left\{\begin{array}{lll} 0,  \ & {\rm if}\; u<\frac{1}{n},\\
 \frac{n}{n-2}x-\frac{1}{n-2},\ &{\rm if}\;  \frac{1}{n}\le u< \frac{n-1}{n},\\
  0,  \ & {\rm if}\; u\ge \frac{n-1}{n}.
 \end{array}
  \right.
\end{eqnarray*}
Then for $i=1,2,\cdots, n$, we have
\begin{eqnarray*}
 {\tilde w}_i=\left\{\begin{array}{llll} \frac{\alpha_1}{n}+\alpha_3,  \ & {\rm if}\; i=1,\\
  \frac{\alpha_1}{n}+\alpha_4+\frac{\alpha_5}{n-2} ,\ &{\rm if}\;  1<i<n, i=k,\\
  \frac{\alpha_1}{n}+\frac{\alpha_5}{n-2},\ &{\rm if}\; 1<i<n, i\neq k,\\
   \frac{\alpha_1}{n}+\alpha_2,\ &{\rm if}\;  i=n,
 \end{array}
  \right.
\end{eqnarray*}
and
$$\widetilde{OWA}_{Q}(a_1,\cdots,a_n)=\alpha_1\frac{1}{n}\sum_{i=1}^n a_i+\alpha_2 a_{(n)}+\alpha_3 a_{(1)}+\alpha_4 a_{(k)}+\alpha_5\frac{1}{n-2}\sum_{i=2}^{n-1}a_{(i)}.$$
In particular, when $\alpha_4=\alpha_5=0$ we recover the results (22) and (23) in Xu (2005); when $\alpha_2=\alpha_4=\alpha_5=0$ and  $\alpha_3=\alpha_4=\alpha_5=0$   we obtain, respectively,  the  {\sl orlike} and {\sl andlike} aggregations in Yager and Filev (1994); when $\alpha_1=\alpha_2=\alpha_3=\alpha_4=0$, we  get the so-called
olympic aggregators in Yager (1998) (where $w_1=w_n=0, w_2=\cdots=w_{n-1}=\frac{1}{n-2}$).

Motivated by  the notion of WOWA (Weighted OWA) in Torra (1997) (see also Llamazares (2016)), we define the
 $\widetilde{WOWA}$ operator which is  a direct extension of $\widetilde{OWA}$.
 Let $p$ and $w$ be two weighting vectors and let $Q$ be a quantifier generating the weighting vector $w$. The
$\widetilde{WOWA}$  operator associated with $p$, $w$ and $Q$ is the function $\widetilde{WOWA}:{\Bbb{R}}^n\rightarrow {\Bbb{R}}$ given by
$$\widetilde{WOWA}_{P,Q}(a_1,\cdots,a_n)=\sum_{i=1}^n {\tilde q}_i a_{(i)},$$
where the weight ${\tilde q}_i$ is defined as:
$${\tilde q}_i=Q\left(1-\sum_{j=1}^{i-1}p_j\right)-Q\left(1-\sum_{j=1}^{i}p_j\right).$$
The $\widetilde{WOWA}$ operator can also be written as:
\begin{equation}
\widetilde{WOWA}_{P,Q}(a_1,\cdots,a_n)=a_{(1)}+\sum_{i=2}^n Q\left(1-\sum_{j=1}^{i-1}p_j\right)(a_{(i)}-a_{(i-1)}).
\end{equation}
In particular, when $p_i\equiv\frac{1}{n}, i=1,2,\cdots,n$,  we get
$$\widetilde{OWA}_{Q}(a_1,\cdots,a_n)=a_{(1)}+\sum_{i=2}^n Q\left(\frac{n+1-i}{n}\right)(a_{(i)}-a_{(i-1)}).$$
The following property was considered for $orness(Q)$ and $WOWA_{P,Q}$ in  Liu and Han (2008).
\begin{theorem} For RIM the quantifiers $Q_1$ and  $Q_2$,     $Q_1(r)\le Q_2(r)$ for every rational point $r\in [0,1]$ if and only if $\widetilde{orness}(Q_1)\ge \widetilde{orness}(Q_2)$, and for  any $n$ aggregated arguments $a_1, a_2, \cdots, a_n$,
$$\widetilde{WOWA}_{P,Q_1}(a_1,\cdots,a_n) \ge \widetilde{WOWA}_{P,Q_2}(a_1,\cdots,a_n).$$
\end{theorem}
\noindent {\bf Proof}. If $Q_1(r)\le Q_2(r)$ for every rational point $r\in [0,1]$, then by using (3.1) one has
$$\widetilde{orness}(Q_1)-\widetilde{orness}(Q_2)=\int_0^1 (Q_2(x)-Q_1(x))dx\ge 0,$$
which is  $\widetilde{orness}(Q_1)\ge \widetilde{orness}(Q_2)$.

Thanks to (3.2), we get
\begin{eqnarray*}
\widetilde{WOWA}_{P,Q_1}(a_1,\cdots,a_n)&-&\widetilde{WOWA}_{P,Q_2}(a_1,\cdots,a_n)= \sum_{i=2}^n Q_1\left(1-\sum_{j=1}^{i-1}p_j\right)(a_{(i)}-a_{(i-1)})\\
&-&\sum_{i=2}^n  Q_2\left(1-\sum_{j=1}^{i-1}p_j\right)(a_{(i)}-a_{(i-1)})\\
&=&\sum_{i=2}^n \left\{Q_1\left(1-\sum_{j=1}^{i-1}p_j\right)- Q_2\left(1-\sum_{j=1}^{i-1}p_j\right)\right\}(a_{(i)}-a_{(i-1)}))\\
&\ge& 0.
\end{eqnarray*}
On the other hand, letting $a_i=1, a_{j}=0, j\neq i$, then from the nonnegativity of
\begin{eqnarray*}
\widetilde{WOWA}_{P,Q_1}(a_1,\cdots,a_n)&-&\widetilde{WOWA}_{P,Q_2}(a_1,\cdots,a_n)=Q_1\left(1-\sum_{j=1}^{i-1}p_j\right)
-Q_1\left(1-\sum_{j=1}^{i}p_j\right) \\
&-& Q_2\left(1-\sum_{j=1}^{i-1}p_j\right)
+Q_2\left(1-\sum_{j=1}^{i}p_j\right),
\end{eqnarray*}
we get the recursive formula for $i=1,2,\cdots,n$:
$$ Q_1\left(1-\sum_{j=1}^{i}p_j\right)-Q_2\left(1-\sum_{j=1}^{i}p_j\right)\le Q_1\left(1-\sum_{j=1}^{i-1}p_j\right)- Q_2\left(1-\sum_{j=1}^{i-1}p_j\right),$$
 from which we get
$$Q_1\left(1-\sum_{j=1}^{i}p_j\right)\le Q_2\left(1-\sum_{j=1}^{i}p_j\right),\; i=1,2,\cdots,n.$$
It follows that $Q_1(r)\le Q_2(r)$ for every rational point $r\in [0,1]$. This completes the proof of Theorem 3.1.

\begin{corollary} If $Q_1(r)\le Q_2(r), r\in [0,1]$, then $\widetilde{orness}(Q_1)\ge \widetilde{orness}(Q_2)$, and for any $n$ aggregated arguments $a_1, a_2, \cdots, a_n$, we have
$$\widetilde{OWA}_{Q_1}(a_1,\cdots,a_n)\ge \widetilde{OWA}_{Q_2}(a_1,\cdots,a_n).$$
\end{corollary}
 Note that the identity quantifier is the smallest concave  RIM quantifier   and also the largest convex RIM quantifier, and when
 $Q(x)=x$, $orness(Q)=\frac12$, $\widetilde{OWA}_{Q}(a_1,\cdots,a_n)=\frac{1}{n}\sum_{i=1}^n a_i$, we can easily get  the following from Corollary 3.1:

\begin{corollary} For any RIM quantifier $Q$, we have

(i)\ if $Q$ is convex function, then $\widetilde{orness}(Q)\ge \frac12$ and $\widetilde{OWA}_{Q}(a_1,\cdots,a_n)\ge  \frac{1}{n}\sum_{i=1}^n a_i$.

(ii)\ if $Q$ is concave function, then $\widetilde{orness}(Q)\le \frac12$ and $\widetilde{OWA}_{Q}(a_1,\cdots,a_n)\le  \frac{1}{n}\sum_{i=1}^n a_i$.
\end{corollary}

\begin{remark} For two RIM quantifiers $Q$ and $G$, the following two sufficient  conditions for $Q\ge G$ were proposed on  generating functions in Liu (2005):

(i)\ $f(s)g(t)\ge f(t)g(s)$, for any $s, t\in [0,1]$ such that $t\ge s$;

(ii)\ $f(s)-f(t)\ge g(s)-g(t)$, for any $s, t\in [0,1]$ such that $t\ge s$;

(iii)\ If $f$ and $g$ is differentiable, $f'(x)\le g'(x), x\in [0,1]$.
\end{remark}

A special class of the OWA operator with
monotonic weights was investigated by Liu and Chen (2004). The following gives the relationship between monotonic weights and the concavity/convexity of RIM quantifier.
Let us start with the standard definitions of convexity and  concavity.

\noindent {\bf Definition 3.2.}\  Let $I$ be an interval in real line $\Bbb{R}$. Then the function $f: I\rightarrow\Bbb{R}$ is said to be convex if
for all $x, y \in I$ and all $\alpha\in [0,1]$, the inequality
$$f(\alpha x+(1-\alpha)y)\le \alpha f(x)+(1-\alpha)f(y)$$
holds. If this inequality is strict for all $x\neq y$ and $\alpha\in (0,1)$, then $f$ is said to be strictly convex.
A closely related concept is that of concavity: $f$ is said to be (strictly) concave if,
and only if, $-f$ is (strictly) convex.

\begin{theorem}  Let $p$ and $w$ be two weighting vectors and let $Q$ be a quantifier generating the weighting vector $w$.  We define
$$q_i=Q\left(\sum_{j=1}^{i}p_j\right)-Q\left(\sum_{j=1}^{i-1}p_j\right), i=1,2,\cdots,n$$
and
$${\tilde q}_i=Q\left(1-\sum_{j=1}^{i-1}p_j\right)-Q\left(1-\sum_{j=1}^{i}p_j\right), i=1,2,\cdots,n.$$
 (i)\; If $Q$ is convex, then $\{q_i\}$ is monotonic  increasing; if $Q$ is concave, then $\{q_i\}$ is monotonic  decreasing.\\
(ii)\; If $Q$ is convex, then $\{\tilde{q}_i\}$ is monotonic  decreasing; if $Q$ is concave, then $\{\tilde{q}_i\}$ is monotonic  increasing.
\end{theorem}
\noindent{\bf Proof}.\; If $Q$ is convex, then
$$q_{i+1}-q_i=Q\left(\sum_{j=1}^{i+1}p_j\right)+Q\left(\sum_{j=1}^{i-1}p_j\right)-2Q\left(\sum_{j=1}^{i}p_j\right)\ge 0,$$
and
$${\tilde q}_{i+1}-{\tilde q}_i=q_{n-i}-q_{n-i+1}\le 0.$$
If $Q$ is concave, then
$$q_{i+1}-q_i=Q\left(\sum_{j=1}^{i+1}p_j\right)+Q\left(\sum_{j=1}^{i-1}p_j\right)-2Q\left(\sum_{j=1}^{i}p_j\right)\le 0,$$
and
$${\tilde q}_{i+1}-{\tilde q}_i=q_{n-i}-q_{n-i+1}\ge 0.$$
This ends the proof of Theorem 3.2.

Letting $p_i\equiv\frac{1}{n}, i=1,2,\cdots,n$,  we get the following corollary.
\begin{corollary}  Let  $w$ be a weighting vector generated  by the quantifier  $Q$. We define
$$w_i=Q\left(\frac{i}{n}\right)-Q\left(\frac{i-1}{n}\right), i=1,2,\cdots,n$$
and
$${\tilde w}_i=Q\left(1-\frac{i-1}{n}\right)-Q\left(1-\frac{i}{n}\right), i=1,2,\cdots,n.$$
 (i)\; If $Q$ is convex, then $\{w_i\}$ is monotonic  increasing; if $Q$ is concave, then $\{w_i\}$ is monotonic  decreasing.\\
(ii)\; If $Q$ is convex, then $\{\tilde{w}_i\}$ is monotonic  decreasing; if $Q$ is concave, then $\{\tilde{w}_i\}$ is monotonic  increasing.
\end{corollary}

\begin{remark} For any  RIM quantifier $Q$, if $Q$ is convex, then   the $OWA_Q$  operator  has the property: small aggregated objects have big weights and big aggregated objects have small weights; if $Q$ is concave, then   the $OWA_Q$  operator  has the property: small aggregated objects have small weights and big aggregated objects have big weights. Contrarily, if $Q$ is convex, then   the ${\widetilde{OWA}}_Q$  operator  has the property: small aggregated objects have small weights and big aggregated objects have big weights; if $Q$ is concave, then   the ${\widetilde{OWA}}_Q$  operator  has the property: small aggregated objects have big weights and big aggregated objects have small weights.
\end{remark}

\vskip 0.2cm
 \section{Elliptical  distributions-based weights-determining  method}
\setcounter{equation}{0}

Xu (2005)  introduced a
procedure for generating the OWA weights based on
 the normal distribution (or Gaussian distribution).  Yager (2007)    referred to
these as Gaussian weights and was described in the following. This is a specific case of the centered OWA operators.
Consider a normal distribution $N(\mu_n, \sigma_n^2)$, where
$$\mu_n=\frac{n+1}{2},\, \sigma_n^2=\frac{1}{n}\sum_{i=1}^n (i-\mu_n)^2=\frac{n^2-1}{12}.$$
The associated OWA weights are defined as:
$$w_i=\frac{e^{-(i-\mu_n)^2/2\sigma_n^2}}{\sum_{j=1}^n e^{-(j-\mu_n)^2/2\sigma_n^2}}, \; i=1,2,\cdots, n.$$
It is clear that $w_i\in [0,1]$ and $\sum_{i=1}^n w_i=1$ and $w_i$ is symmetric, that is $w_i=w_{n+1-i}$.

The Xu's method on the normal type OWA weighting
vector inspires us to consider a more general class of
OWA aggregation operators of this type. We shall refer
to these as elliptical OWA operators.

\noindent{\bf Definition 4.1.}\; Let $X$ be the continuous random variable,  we say $X$ belonging to the class of elliptical distributions
if its density can be expressed as:
$$f_X(x)=\frac{C}{\sigma}g\left[\left(\frac{x-\mu}{\sigma}\right)^2\right],\;-\infty<x<\infty,$$
 for some so-called density generator $g$ (which is a function of non-negative variables)  satisfying the condition:
 $$0<\int_0^{\infty}x^{-\frac12}g(x)dx<\infty,$$
 and a normalizing constant $C$ given by
 $$C=\left[\int_0^{\infty}x^{-\frac12}g(x)dx\right]^{-1}.$$

For the  normal distribution $N(\mu,\sigma^2)$,
  it is straightforward to show that its density generator has  the form $g(x)=e^{-\frac{x}{2}}.$ In general, elliptical
distributions can be bounded or unbounded, unimodal or multimodal, the class of
elliptical distributions consists of the class of symmetric distributions.
\begin{table}\caption{Some known elliptical distributions with their density generators}
\centering
\begin{tabular}{|c||c|}
\hline  Family & Density generators $g(u)$\\
\hline
 Cauchy &  $g(u)=\frac{1}{1+u}$\\
 Exponential Power& $g(u)=e^{-r u^s}, r, s>0$\\
 Laplace &$g(u)=e^{-|u|}$\\
 Logistic& $g(u)=\frac{e^{-u}}{(1+e^{-u})^2}$\\
 Normal &$g(u)=e^{-u/2}$\\
 Student-$t$ & $g(u)=\left(1+\frac{u}{m}\right)^{{1+m}/2}, m>0$ an integer\\
\hline
\end{tabular}
\end{table}
For details, see Landsman and Valdez (2003).

Following Xu (2005), we define
$$\mu_n=\frac{1+n}{2},\; \;\; \sigma_n^2=\frac{1}{n}\sum_{i=1}^n (i-\mu_n)^2,$$
and
\begin{equation}
w_i=\frac{g\left[\left(\frac{i-\mu_n}{\sigma_n}\right)^2\right]}{\sum_{j=1}^n g\left[\left(\frac{j-\mu_n}{\sigma_n}\right)^2\right]}, \;\;\; i=1,2,\cdots,n.
\end{equation}
It can be shown that these weights satisfy the conditions  $w_i\in [0,1]$ and $\sum_{i=1}^n w_i=1$.

Similar to Theorem 2 in Xu (2005), we have
\begin{theorem}
(1) The weights $w_i, i=1,2,\cdots,n$ are symmetrical, that is,
\begin{equation}
w_i=w_{n+1-i},\;\;\; i=1,2\cdots,n.
\end{equation}
(2) If $g$ is non-increasing, then $w_i\le w_{i+1}$ for all $i=1,2,\cdots, [\frac{1+n}{2}]$, and $w_i\ge w_{i+1}$ for all $i=[\frac{1+n}{2}]+1,\cdots, n,$
where $[\cdot]$ is the usual round operation.\\
(3) If $n$ is odd, then the weight $w_i$
reaches its maximum when $i=[\frac{1+n}{2}]$; if $n$ is even, then the weight $w_i$
reaches its maximum when  $i=[\frac{1+n}{2}]$ or $i=[\frac{1+n}{2}]+1$.\\
(4) Orness(w)=0.5.
\end{theorem}
\noindent{\bf Proof}.\ (1) It follows that (4.1) that
 \begin{eqnarray*}
w_{n+1-i}&=&\frac{g\left[\left(\frac{n+1-i-\mu_n}{\sigma_n}\right)^2\right]}{\sum\limits_{j=1, j\neq i}^n g\left[\left(\frac{j-\mu_n}{\sigma_n}\right)^2\right]+ g\left[\left(\frac{n+1-i-\mu_n}{\sigma_n}\right)^2\right]}\\
&=&\frac{g\left[\left(\frac{i-\mu_n}{\sigma_n}\right)^2\right]}{\sum\limits_{j=1, j\neq i}^n g\left[\left(\frac{j-\mu_n}{\sigma_n}\right)^2\right]+ g\left[\left(\frac{i-\mu_n}{\sigma_n}\right)^2\right]}\\
&=&\frac{g\left[\left(\frac{i-\mu_n}{\sigma_n}\right)^2\right]}{\sum\limits_{j=1}^n g\left[\left(\frac{j-\mu_n}{\sigma_n}\right)^2\right]}=w_i, \;\;\; i=1,2,\cdots,n.
 \end{eqnarray*}
(2)  Because
$$\left(\frac{i-\mu_n}{\sigma_n}\right)^2>\left(\frac{i+1-\mu_n}{\sigma_n}\right)^2\; {\rm for}\; i=1,2,\cdots, \left[\frac{1+n}{2}\right],$$
and
$$\left(\frac{i-\mu_n}{\sigma_n}\right)^2<\left(\frac{i+1-\mu_n}{\sigma_n}\right)^2\; {\rm for}\; i=\left[\frac{1+n}{2}\right]+1,\cdots, n,$$
and note that $g$ is non-increasing,
then
$$g\left(\left(\frac{i-\mu_n}{\sigma_n}\right)^2\right)\le g\left(\left(\frac{i+1-\mu_n}{\sigma_n}\right)^2\right), i=1,2,\cdots, \left[\frac{1+n}{2}\right],$$
and
$$g\left(\left(\frac{i-\mu_n}{\sigma_n}\right)^2\right)\ge g\left(\left(\frac{i+1-\mu_n}{\sigma_n}\right)^2\right), \; i=\left[\frac{1+n}{2}\right]+1,\cdots, n.$$
The result follows since the function $f(x)=x/(a+x), x>0$ is increasing function, where $a>0$ is a constant.

(3) Obvious.

(4)  This result can be derived directly from Theorem in Yager (2007) and (4.2).
 This ends the proof of Theorem 4.1.

As pointed out by  Su et al. (2016), the above method is simple and straightforward,  but sometimes there exists some
problems in actual applications. For example,  the same value may have  different weights
 because of their different positions; For details see Su et al. (2016).
In order to avoid the issue above, Wang and Xu (2008) introduced a weighting method
for the weighted arithmetic aggregation (WAA)  operator, which is based on the normal probability density function and
the given arguments simultaneously:
Given a collection of n preference values $a_j (j = 1, 2, \cdots, n)$, and let $w =
(w_1,w_2, \cdots, w_n)$ be the weighting vector of the WAA operator, then Wang and Xu
(2008) gave the following formula:
\begin{equation}
w_i=\frac{e^{-(a_i-\mu)^2/2\sigma^2}}{\sum_{j=1}^n e^{-(a_j-\mu)^2/2\sigma^2}}, \; i=1,2,\cdots, n,
\end{equation}
 where $\mu$ is the mean of the collection of $a_j (j = 1, 2, \cdots, n)$, and $\sigma$ is the standard
deviation of   $a_j ( j = 1, 2, \cdots, n)$, i.e.,
$$\mu=\frac{\sum_{j=1}^n a_j}{n},\;\; \sigma=\sqrt{\frac{1}{n}\sum_{j=1}^n(a_j-\mu)^2}.$$
Motivated by  the method above, for any density generator $g$, we define
\begin{equation}
w_i=\frac{g\left[\left(\frac{a_i-\mu}{\sigma}\right)^2\right]}{\sum_{j=1}^n g\left[\left(\frac{a_j-\mu}{\sigma}\right)^2\right]}, \;\;\; i=1,2,\cdots,n.
\end{equation}
It can be shown that these weights satisfy the conditions  $w_i\in [0,1]$ and $\sum_{i=1}^n w_i=1$.
In particular, (4.4) become (4.3) whenever $g(x)=e^{-\frac{x}{2}}$.

The following theorem shows that  $OWA_{P,Q}$ and $\widetilde{WOWA}_{P,Q}$ are the same for some cases:
\begin{theorem} Let $g$ be an elliptical density function  which is symmetric about 0.5, let $p=(p_1,\cdots, p_n)$  be a weighting vectors.  Assume that  $Q$ is a   RIM quantifier
generated by $g$, that is
$$Q(x)=\frac{1}{\int_0^1 g(x)dx}\int_0^x g(y)dy, \;\;x\in [0,1].$$
 We define $${\tilde q}_i=Q\left(1-\sum_{j=1}^{i-1}p_j\right)-Q\left(1-\sum_{j=1}^{i} p_j\right),\; i=1, 2,\cdots, n,$$
 and
 $$q_i=Q\left(\sum_{j=1}^{i} p_j\right)-Q\left(\sum_{j=1}^{i-1}p_j\right),\; i=1, 2,\cdots, n.$$
 Then  $$w_i={\tilde w}_i, \; i=1, 2,\cdots, n.$$
\end{theorem}
\noindent{\bf Proof.}\,  Since $g$ is   symmetric about $\frac12$, it follows that $g(\frac12+x)=g(\frac12-x)$ for any $x$. We have
\begin{eqnarray*}
Q\left(\frac12+x\right)&+&Q\left(\frac12-x\right)\\
&=&\frac{1}{\int_0^1 g(x)dx}\int_0^{x+\frac12} g(y)dy+\frac{1}{\int_0^1 g(x)dx}\int_0^{\frac12-x} g(y)dy\\
&=&\frac{1}{\int_0^1 g(x)dx}\int_{-\frac12}^{x} g\left(\frac12+y\right)dy+\frac{1}{\int_0^1 g(x)dx}\int_{-\frac12}^{-x} g\left(y+\frac12\right)dy\\
&=&\frac{1}{\int_0^1 g(x)dx}\int_{-\frac12}^{x} g\left(\frac12+y\right)dy-\frac{1}{\int_0^1 g(x)dx}\int_{\frac12}^{x} g\left(\frac12-y\right)dy\\
&=&\frac{1}{\int_0^1 g(x)dx}\int_{-\frac12}^{\frac12} g\left(\frac12+y\right)dy\\
&=&\frac{1}{\int_0^1 g(x)dx}\int_{0}^{1} g(y)dy=1,\\
\end{eqnarray*}
 from which we get
\begin{eqnarray*}
q_i&=&1-Q\left(\sum_{j=1}^{i-1}p_j\right)+Q\left(\sum_{j=1}^{i}p_j\right)-1\\
&=&Q\left(1-\sum_{j=1}^{i-1}p_j\right)-Q\left(1- \sum_{j=1}^{i}p_j\right)\\
&=&{\tilde q}_i.
\end{eqnarray*}
This ends the proof of Theorem 4.2.

Letting $p_1=\cdots=p_n=\frac{1}{n}$ in Theorem 4.2, we get
\begin{corollary} Let $g$ be an elliptical density function  which is symmetric about 0.5.  Assume that  $Q$ is a   RIM quantifier
generated by $g$, that is
$$Q(x)=\frac{1}{\int_0^1 g(x)dx}\int_0^x g(y)dy, \;\;x\in [0,1].$$
We define $${\tilde w}_i=Q\left(1-\frac{i-1}{n}\right)-Q\left(1-\frac{i}{n}\right),\; i=1, 2,\cdots, n,$$
 and
 $$w_i=Q\left(\frac{i}{n}\right)-Q\left(\frac{i-1}{n}\right),\; i=1, 2,\cdots, n.$$
 Then  $$w_i={\tilde w}_i, \; i=1, 2,\cdots, n.$$
\end{corollary}

The centered OWA operators were  introduced by Yager (2007).
 An OWA operator is said to be a centered OWA operator if its associated weighting
vector $W$ satisfies the following conditions: $w_i=w_{n+1-i}, i=1,2\cdots, [\frac{n}{2}];$ $w_i<w_j,$ whenever $i<j\le [\frac{n+1}{2}]$, or
$i>j\ge [\frac{n+1}{2}]$; $w_i>0$ for any $i$. Notice that the centered OWA operators allow us to give more importance to the central values and less
weight to the extreme scores. Specific cases of the centered OWA operators  can be found in Llamazares (2018).
One  important method to construct  the weights associated with the centered OWA operators is by using the  centering functions  introduced by Yager (2007).  A function $g: [0, 1] \to {\Bbb R}^+$ is called a centering
function if it is  symmetric,  unimodal and satisfying condition
$$0<\int_0^1 g(x)dx<\infty.$$
\begin{theorem}(Yager (2007))\; Let $g$ be a centering function. If $f$ is a function
defined on the unit interval such that
$$f(x)=\frac{1}{\int_0^1 g(x)dx}\int_0^x g(y)dy, \;\;x\in [0,1],$$
then $f$ is a BUM function (an acronym for Basic Unit
interval Monotonic function) that
generates centered weights.
\end{theorem}
We remark that the domain of $g$ is not necessary $[0,1]$ as shows in the follwing:
\begin{theorem}\;   Let $g$ be an  unimodal elliptical density function  which is symmetric about 0.5.  We define
$$Q(x)=K\int_0^x g(y)dy, \;\;x\in [0,1],$$
where $$K=\frac{1}{\int_0^1 g(x)dx}.$$
  Then $f$ is a BUM function   that generates the centered weights.
\end{theorem}
\noindent{\bf Proof}.\,  First we show that $Q$ is a BUM function. Obviously, $Q(0)=0, Q(1)=1$ and
for $x_2>x_1$,
$$Q(x_2)-Q(x_1)=\frac{1}{\int_0^1 g(x)dx}\int_{x_1}^{x_2} g(y)dy\ge 0,$$
sine $g\ge 0$.

Now we  show that the weights generated from this
$Q$ satisfy the conditions of a centered weighting vector
for all $n$, (i)\;$w_i=w_{n+1-i}, i=1,2\cdots, [\frac{n}{2}];$ (ii)\; $w_i<w_j,$ whenever $i<j\le [\frac{n+1}{2}]$, or
$i>j\ge [\frac{n+1}{2}]$; (iii)\; $w_i>0$ for any $i$.

(i) Symmetry: Because of $Q\left(\frac12+x\right)+Q\left(\frac12-x\right)=1$, we get
\begin{eqnarray*}
w_{n+1-i}&=&Q\left(\frac{n+1-i}{n}\right)-Q\left(\frac{n-i}{n}\right)\\
&=&Q\left(\frac12+\frac12-\frac{i-1}{n}\right)-Q\left(\frac12+\frac12-\frac{i}{n}\right)\\
&=&Q\left(\frac{i}{n}\right)-Q\left(\frac{i-1}{n}\right)\\
&=&w_i.
\end{eqnarray*}
(ii) Unimodality: For $i<j\le [\frac{n+1}{2}]$,
$$w_j-w_i=K\int_0^{\frac1n}\left\{g\left(z+\frac{j-1}{n}\right)-g\left(z+\frac{i-1}{n}\right)\right\}\ge 0$$
since $g\left(z+\frac{j-1}{n}\right)\ge g\left(z+\frac{i-1}{n}\right)$. Thus $w_j> w_i$.

For $i>j\ge [\frac{n+1}{2}]$,
$$w_j-w_i=K\int_0^{\frac1n}\left\{g\left(z+\frac{j-1}{n}\right)-g\left(z+\frac{i-1}{n}\right)\right\}\ge 0$$
since $g\left(z+\frac{j-1}{n}\right)\ge g\left(z+\frac{i-1}{n}\right)$. Thus $w_j> w_i$.

(iii)\; Inclusiveness: Since $g> 0$, we have
$$w_i=Q\left(\frac{i}{n}\right)-Q\left(\frac{i-1}{n}\right)=K\int_{\frac{i-1}{n}}^{\frac{i}{n}}g(x)dx>0.$$
This ends the proof of Theorem 4.4.

\section{ Concluding remarks}

In this paper, we have surveyed the existing main steps for determining
the OWA weights. We  introduced a new method to determine  weights by the quantifier function and compare the properties with usual weights determined by the same quantifier function, the associated operators are call the dual
OWA operators.  Based on the elliptical distribution, we have developed a novel
practical method for obtaining the weight vector of the OWA operator. Some of its desirable properties  have been investigated in detail.
The key characteristic of the elliptical OWA operators, as the  normal type OWA  operators,
 is that it can relieve the influence of unfair arguments
on the decision results by assigning low weights to those ``false" or ``biased"
ones.

\noindent{\bf Acknowledgements.} \ 
The research   was supported by the National
Natural Science Foundation of China (No. 11171179, 11571198, 61273209).

\end{document}